\documentclass[11pt]{article}

\usepackage{amsmath,amsfonts,bm}

\def\eqref#1{equation~\ref{#1}}

\def\1{\bm{1}}

\DeclareMathAlphabet{\mathsfit}{\encodingdefault}{\sfdefault}{m}{sl}
\SetMathAlphabet{\mathsfit}{bold}{\encodingdefault}{\sfdefault}{bx}{n}

\usepackage{graphicx}
\usepackage{xcolor}
\definecolor{myviolet}{RGB}{128,0,128}
\usepackage[colorlinks=true,citecolor=myviolet,linkcolor=cyan]{hyperref}       
\usepackage{url}
\usepackage{wrapfig}
\usepackage{xspace}
\usepackage{subcaption}
\usepackage{wrapfig}
\usepackage[table]{xcolor} 
\definecolor{lightgray}{HTML}{EFEFEF}
\usepackage{amssymb} 
\usepackage{pifont}  
\usepackage{booktabs} 
\usepackage{fontawesome}  
\newcommand{\xmark}{\textcolor{brown}{\ding{55}}}  
\newcommand{\cmark}{\textcolor{teal}{\checkmark}}
\newcommand{\ourbench}{\textsc{StockBench}\xspace}
\usepackage{lipsum}
\usepackage{enumitem}
\usepackage{amsmath}

\usepackage{graphicx} 
\usepackage{booktabs} 
\usepackage{siunitx}  
\usepackage{caption}  
\usepackage{multirow} 

\usepackage{tikz}
\usepackage{pgfplots}
\pgfplotsset{compat=1.18}

\title{\ourbench:  Can LLM Agents Trade Stocks Profitably In \\Real-world Markets?}

\author{
Yanxu Chen$^{\spadesuit\heartsuit}$\thanks{Equal contribution.}\thanks{Work was completed during an internship at Tsinghua University.}
\quad
Zijun Yao$^{\spadesuit}$\footnotemark[1]
\quad
Yantao Liu$^{\spadesuit}$\footnotemark[1]
\quad
Amy Xin$^{\spadesuit}$ \\
\textbf{Jin Ye}$^{\heartsuit}$
\quad
\textbf{Jianing Yu}$^{\heartsuit}$
\quad
\textbf{Lei Hou}$^{\spadesuit}$
\quad
\textbf{Juanzi Li}$^{\spadesuit}$ \vspace{0.05in}\\
$^{\spadesuit}$Tsinghua University
\qquad
$^{\heartsuit}$Beijing University of Posts and Telecommunications \vspace{0.05in}\\
\faEnvelope\ \texttt{yaozj20@mails.tsinghua.edu.cn, cyx666@bupt.edu.cn} \vspace{0.05in}\\
{\small  \faFolderOpen\ \textbf{Project:} \href{https://stockbench.github.io/}{\texttt{https://stockbench.github.io/}}} \\[-0.1em]
{\small  \faCode\ \textbf{GitHub:} \href{https://github.com/ChenYXxxx/stockbench}{\texttt{https://github.com/ChenYXxxx/stockbench}}} \\
}

\usepackage[preprint]{acl}

\usepackage{times}
\usepackage{latexsym}

\usepackage[T1]{fontenc}

\usepackage[utf8]{inputenc}

\usepackage{microtype}

\usepackage{inconsolata}

\usepackage{graphicx}

\begin{document}
\maketitle

\begin{abstract}
    Large language models (LLMs) demonstrate strong potential as autonomous agents, with promising capabilities in reasoning, tool use, and sequential decision-making. 
    While prior benchmarks have evaluated LLM agents in various domains, the financial domain remains underexplored, despite its significant economic value and complex reasoning requirements. 
    Most existing financial benchmarks focus on static question-answering, failing to capture the dynamics of real-market trading. 
    To address this gap, we introduce \ourbench, a contamination-free benchmark designed to evaluate LLM agents in realistic, multi-month stock trading environments.
    Agents receive daily market signals---including prices, fundamentals, and news---and make sequential buy, sell, or hold decisions. 
    Performance is measured using financial metrics such as cumulative return, maximum drawdown, and the Sortino ratio, capturing both profitability and risk management.
    We evaluate a wide range of state-of-the-art proprietary and open-source LLMs.
    Surprisingly, most models struggle to outperform the simple buy-and-hold baseline, while some models demonstrate the potential to achieve higher returns and stronger risk management. 
    These findings highlight both the challenges and opportunities of LLM-based trading agents, showing that strong performance on static financial question-answering do not necessarily translate into effective trading behavior.
    We release \ourbench as an open-source benchmark to enable future research on LLM-driven financial agents.
\end{abstract}


\section{Introduction}
\label{sec:intro}

Large language models (LLMs) have enabled extensive exploration of autonomous agents, demonstrating strong capabilities in reasoning, tool use, and long-horizon decision making~\citep{gpt4o, claude3.7, gemini2.5, deepseekv3, r1, llama4, qwen2.5, qwen2.5vl, o3}.
These agentic capabilities have been verified by benchmarks in various domains, such as software engineering~\citep{jimenez2024swebench,yang2024sweagent}, scientific discovery~\citep{astabench}, and marketing~\citep{chen2025xbench,barres2025tau2bench}.
Evaluations using state-of-the-art LLMs such as GPT-5~\citep{gpt5} and Claude-4~\citep{claude4}, highlight the protential of LLM agents to support workflow automation and productivity gains.
As LLMs continue to improve, their growing agentic capabilities increasingly push applications toward real-world deployment with tangible economic value.

Among various agent application domains, the financial domain stands out with its direct link to economic value and the high stakes of decision making~\citep{wu2023bloomberggpt,lee2024finllms,nie2024survey}.
To rigorously evaluate the profitability and risk-management capabilities of LLM agents in financial settings, an ideal benchmark should satisfy three key principles:
\textbf{(1)~Realistic Market Interaction.} Agents should operate in a dynamic market environment, reacting to real-time price movements and news events.
\textbf{(2)~Continuous Decision Making.} Agents should make sequential trading decisions over extended horizons, reflecting the iterative nature of real investment strategies.
\textbf{(3)~Contamination-Free Data.} To ensure fair and reliable evaluation, agents must not have prior exposure to the test data during training, requiring careful data curation and strict temporal separation.

\begin{table*}[t]
    \centering
    \resizebox{\textwidth}{!}{
    \begin{tabular}{l|c|c|c|c|c}
    \toprule
    \textbf{Benchmark} & \textbf{Market} & \textbf{Multi Month} & \textbf{Continuous} & \textbf{Contamination} & \textbf{Direct Economic} \\
    & \textbf{Simulation} & \textbf{Horizon} & \textbf{Decision} & \textbf{Free} & \textbf{Value} \\
    \midrule
    FinQA~\citep{chen2021finqa}              & \xmark & \xmark & \xmark & \xmark & \xmark \\
    ConvFinQA~\citep{convfinqa}              & \xmark & \xmark & \xmark & \xmark & \xmark \\
    FLUE~\citep{flue}                        & \xmark & \xmark & \xmark & \xmark & \xmark \\
    FinEval~\citep{fineval}                  & \xmark & \xmark & \xmark & \xmark & \xmark \\
    CPA-QKA~\citep{scores2skills}            & \xmark & \xmark & \xmark & \xmark & \xmark \\
    BizFinBench~\citep{bizfinbench}         & \xmark & \xmark & \xmark & \xmark & \xmark \\
    Finance Agent Benchmark~\citep{financeagent} & \cmark & \xmark & \cmark & \xmark & \xmark \\
    INVESTORBENCH~\citep{investorbench}     & \cmark & \cmark & \cmark & \xmark & \cmark \\
    FinSearchComp~\citep{finsearchcomp}     & \xmark & \cmark & \cmark & \xmark & \cmark \\
    \midrule
    \textbf{\ourbench (Ours)}               & \cmark & \cmark & \cmark & \cmark & \cmark \\
    \bottomrule
    \end{tabular}
    }
    \caption{Comparison of \ourbench with existing financial benchmarks.}
    \label{tab:finsearchcomp}
\end{table*}

However, existing financial benchmarks for LLM agents largely focus on static question-answering tasks~\citep{chen2021finqa, zhu-etal-2021-tat, yin2023finbench}, which are designed to test the financial knowledge of LLMs but fail to capture the dynamics of realistic trading scenarios.
Although recent efforts like INVESTORBENCH~\citep{li-etal-2025-investorbench} take a step towards simulating trading environments, they are limited to single-stock-trading and rely on historical data prior to 2021, raising concerns about data contamination and outdated market conditions.

To bridge this gap, we propose \ourbench, an evolving benchmark that places LLM agents in realistic stock-trading environments, directly evaluating their profitability and risk-management capabilities. Specifically, \ourbench is designed to be:  
\textbf{(1)~Realistic.} Agents receive daily market signals including prices, company fundamentals, and news headlines, reflecting real-world trading conditions.  
\textbf{(2)~Continuous.} Agents make sequential daily trading decisions (buy, sell, or hold) over a multi-month horizon, mirroring the iterative nature and long-term nature of investment strategies.  
\textbf{(3)~Contamination-Free.} The benchmark is instantiated using recent market data from March 2025 to July 2025 and will be continuously updated to prevent overlap with the training corpora of LLMs.
Performance is evaluated using key financial metrics such as cumulative return, maximum drawdown, and the Sortino ratio, providing a quantitative assessment of both profitability and risk control.

As a proof of concept, we evaluate a diverse set of LLM agents, including both proprietary models (\textit{e.g.,} GPT-5~\citep{gpt5}, Claude-4~\citep{claude4}) and open-weight models (\textit{e.g.,} Qwen3~\citep{qwen3}, Kimi-K2~\citep{kimi-k2}, GLM-4.5~\citep{glm4.5}), alongside an equal-weight buy-and-hold baseline.
Surprisingly, despite their strong performance on financial QA benchmarks, most LLM agents fail to outperform this simple baseline in terms of both cumulative return and risk-adjusted return.
This finding suggests that success on static financial QA does not necessarily translate into effective trading strategies in dynamic market environments, underscoring a key challenge for LLM-based financial agents.

The main contributions of this work are summarized as follows:
\begin{itemize}[leftmargin=12pt, itemsep=0pt, topsep=4pt]
    \item We introduce \ourbench, a novel benchmark for evaluating LLM agents in a realistic stock-trading environment, measuring both profitability and risk-management capabilities.
    \item We propose a comprehensive evaluation framework that incorporates realistic market dynamics, diverse input data, and multiple financial metrics to holistically assess agent performance.
    \item We conduct extensive experiments across a range of state-of-the-art LLMs, revealing their current limitations in achieving profitable trading strategies and underscoring the need for further methodological advances.
    \item We open-source the implementation of \ourbench to facilitate reproducibility and to encourage community contributions, fostering further research on LLM-based financial agents.
\end{itemize}

\section{\ourbench}



\begin{figure*}[t]
    \centering
    \includegraphics[width=1\textwidth]{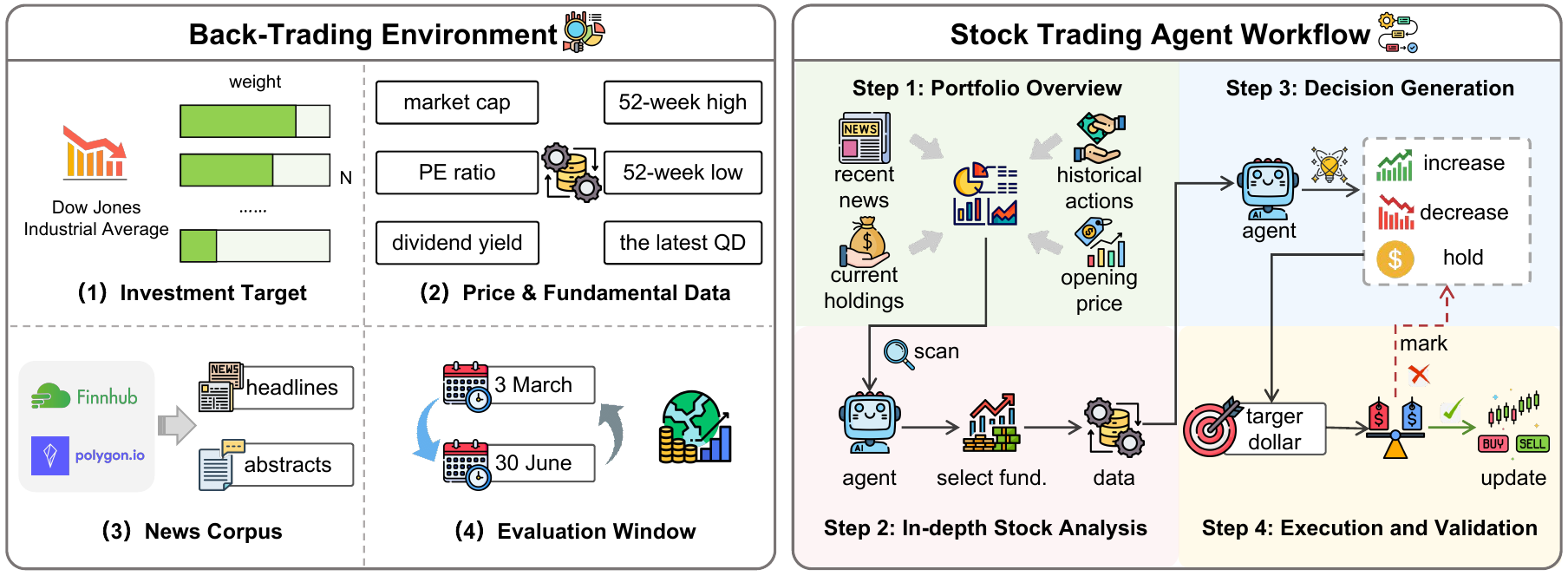}
    \caption{
Overview of \ourbench. The design of \ourbench includes a back-trading benchmark dataset, and an associated workflow that converts backbone LLMs into agents.
}
    \label{fig:pdfimage}
\end{figure*}

The construction of \ourbench consists of two main building blocks.
(1) A back-trading environment, which contains historical data necessary for stock-trading decision making.
We simulate real-world stock trading using this back-trading setup.
(2) An associated  stock-trading agent workflow.
This workflow allows us to evaluate LLM backbones as agents to engage in the back-trading environment.
The overall framework of \ourbench is demonstrated in Figure~\ref{fig:pdfimage}.



\subsection{Back-Trading Environment}


We design the back-trading environment to simulate realistic stock trading, where trading agents are exposed only to data available up to the time of each decision.
To set up the environment, we identify three critical sources of information for trading decision making:
(1)~A bundle of investment targets, which defines the scope of the environment. 
We pre-define these investment targets to facilitate reproducibility of the evaluation on \ourbench.
(2)~Historical market data, which includes both the prices and fundamental indicators.
These enable the evaluated trading agents to perform quantitative analysis.
(3)~News corpora, which capture events that drive stock price fluctuations.




\paragraph{Investment Targets.}  
The investment targets are a bundle of stocks that allow the trading agents to perform buy and sell operations.
We manually select the investment targets 
to prevent potential outcome fluctuations caused by stock selection---\textit{e.g.,} trading agents might otherwise happen to pick a stock driven by irrational market sentiment---thereby stabilizing the evaluation results.

\begin{figure}[!t]
    \centering
    \includegraphics[width=1\linewidth]{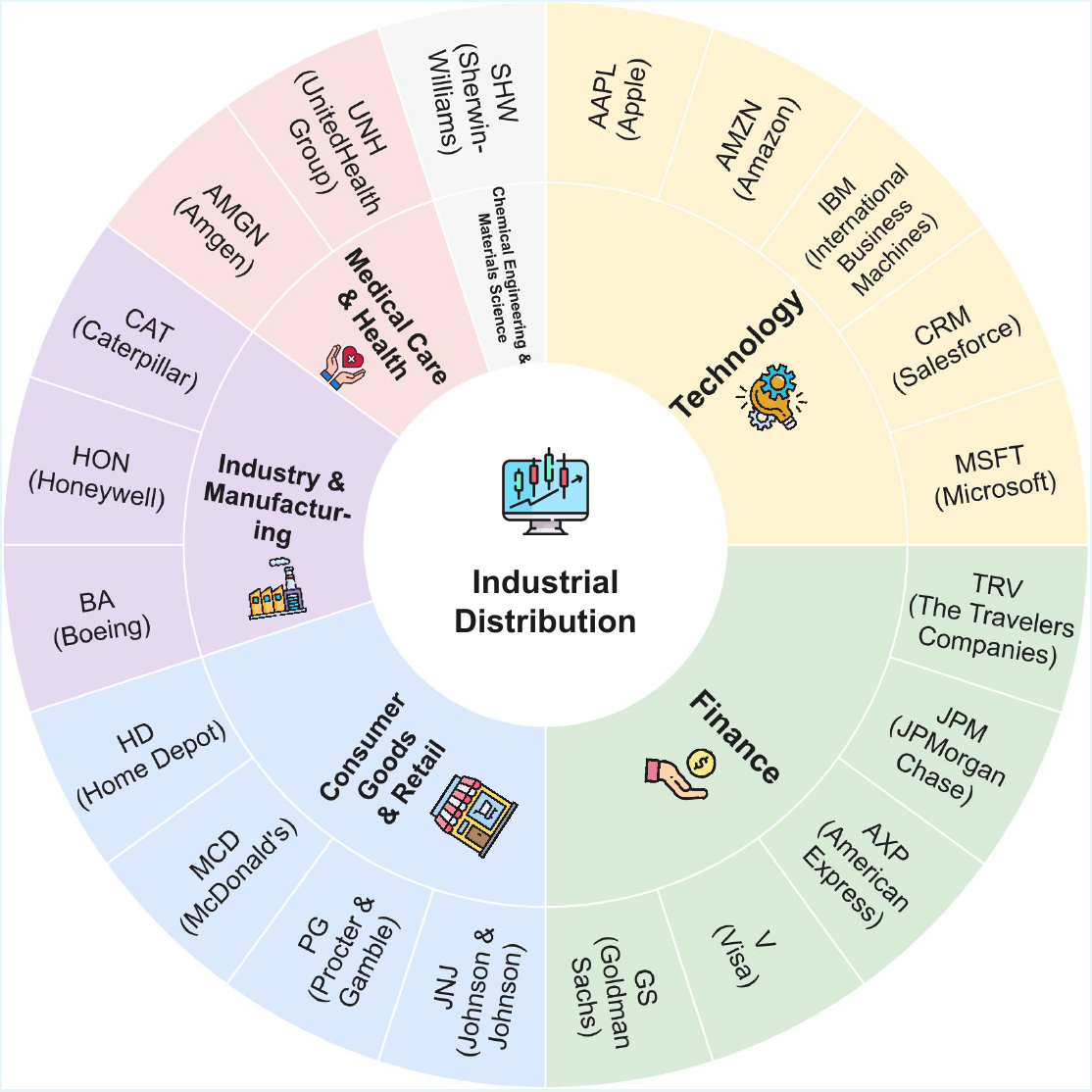}
    \caption{Industry distribution of selected stocks.}
    \label{fig:stock_fig}
\end{figure}

To this end, we select $20$ stocks from the Dow Jones Industrial Average (DJIA) with the highest weights as our investment targets.
In particular, high-weighted DJIA stocks are representative of the global stock market and are less prone to short-term irrational sentiment-driven events.
Constraining the trading action space to our selected investment targets mirrors real-world investor attention while keeping the dataset computationally tractable.
Moreover, information about these well-known stocks is transparent and easy to collect, being readily accessible through web search engines.
We show the distribution of the selected investment targets across different industries in Figure~\ref{fig:stock_fig}.
Our selection covers technology, finance, and manufacturing, ensuring stock diversity.



\paragraph{Historical Market Data.}  
We collect and preserve historical market data containing key quantitative information.
For each stock, we use official opening prices
with a concise set of fundamental metrics such as market capitalization, price-to-earnings (P/E) ratio, dividend yield, and trading range. 
These signals provide a reliable snapshot of company health and valuation, supporting informed decision making.
We also retain the timestamps of the collected data to prevent 
leakage of future information to the agent.


\paragraph{News Corpora.}  
We construct news corpora for stocks to enable stock-trading agents to interpret both sentiments and events in a manner that resembles how retail investors react to market narratives.
For each stock, we collect news articles released within the previous $48$ hours on a daily basis.
These articles are retrieved using news-search API\footnote{\url{https://finnhub.io/}} with time restrictions.
Since news analysis consumes substantial context length in backbone LLMs, we balance information coverage and computational cost by preserving the top five relevant news articles each time the search engine returns results.

We also carefully select the time window for collecting data in the back-trading environment.
In principle, the evaluation window should satisfy two conditions: (1) the included stock information must not have been exposed to the evaluated stock-trading agents during their model training stages; and (2) the window should be sufficiently long to mitigate the impact of random noise that affects only short periods of time.
To this end, we collect data spanning from \texttt{March 3, 2025} to \texttt{June 30, 2025}, a four-month period that includes both volatility and trend reversals.
This period also falls after the knowledge cutoff of mainstream LLMs, ensuring no data leakage.
It is worth noting that we will continuously update the back-trading environment to avoid overlap with the training corpora of contemporary LLMs.


\subsection{Stock-trading Agent Workflow}

We provide a stock-trading agent workflow that enables backbone LLMs to interact with the back-trading environment as agents.
The design of the workflow follows two goals. 
(1) Minimal workflow. 
We keep the workflow minimal, since overly complicated workflows introduce inductive biases that may favor certain backbone LLMs.
(2) Realistic. We design the workflow to align with the iterative decision-making process of retail investors.

In particular, we follow previous frameworks~\citep{zhang2020deep,tsantekidis2017forecasting,moody2001learning,deng2016deep} and organize the stock-trading workflow into four essential stages: portfolio overview, in-depth stock analysis, decision generation, and execution and validation.

Overall, the design prioritizes realism, fairness, and reproducibility, in line with earlier studies on benchmark construction for trading environments.

\paragraph{Step 1: Portfolio Overview.}  
The agent first scans all available stocks in the market (the ``investment target''), receiving relevant data for each stock. 
This includes recent news, current holdings of the agent, historical actions, and the opening price. 
This step mirrors how a trader assesses the broader market and the overall status of each stock in their portfolio.

\paragraph{Step 2: In-Depth Stock Analysis.}  
After the initial overview, the agent selects specific stocks for deeper analysis. 
For these selected stocks, the agent is provided with additional fundamental data such as market capitalization, P/E ratio, and dividend yield. 
This step simulates how a trader focuses on a subset of stocks identified in the initial overview, examining their financial health and other key metrics in greater depth.

\paragraph{Step 3: Decision Generation.}  
With the enriched context, the agent generates decisions for each stock, choosing between three actions: 
(1) increase, (2) decrease, or (3) hold the position. 
These options ensure that 
the agent's actions  
are clear, actionable, and executable within the constraints of a retail investor's decision making process.

\paragraph{Step 4: Execution and Validation.}  
Finally, the decisions are executed by converting dollar targets into share quantities based on the opening price. 
If the decisions of the agents exceed available liquidity, the system flags the issue and requires the agent to revise its decisions until they can be executed within available resources. 
Once validated, the new portfolio weights are locked, and the simulation advances to the next day.

\subsection{Features of \ourbench}

We now discuss how the design of \ourbench satisfies the following key principles:


    \paragraph{Realistic Market Interaction.}   
    The design of the back-trading environment mimics real-world trading scenarios through three key elements: (1) a carefully selected bundle of investment targets, (2) reliable price and fundamental data, and (3) a concise yet timely news corpus.  
    These elements ensure that the agent is exposed to information mirroring the complexities of real trading environments, 
    avoiding unrealistic or overly expansive inputs.  

    \paragraph{Continuous Decision Making.}   
    In the workflow, the agent first performs a portfolio overview, then conducts in-depth stock analysis, and finally generates daily trading decisions (buy, sell, or hold) based on this analysis. 
    These steps reflect the continuous decision-making process of retail investors, enabling the agent to adapt its strategies over time in response to market conditions.

    \paragraph{Contamination-Free Data.}   
    We ensure that the agent has no prior exposure to the test data during its training. 
    To achieve this, the benchmark is instantiated using recent market data, ensuring temporal separation and avoiding any overlap with the training corpora of contemporary LLMs.

\begin{table}[t] 
    \centering
    \setlength{\tabcolsep}{2.0pt}
    \renewcommand{\arraystretch}{0.85}
    \rowcolors{2}{white}{lightgray}
    \scalebox{0.97}{
    \begin{tabular}{lrrrrrr}
    \toprule
    \textbf{Model} & \textbf{RT} & \textbf{DDN} & \textbf{Sortino} & \textbf{Rank} \\
    \midrule
    Kimi-K2 & $1.9$ & $-11.8$ & $\mathbf{0.0420}$ & $1$ \\
    Qwen3-235B-Ins & $2.4$ & $\mathbf{-11.2}$ & $0.0299$ & $2$ \\
    GLM-4.5 & $2.3$ & $-13.7$ & $0.0295$ & $3$ \\
    Qwen3-235B-Think & $\textbf{2.5}$ & $-14.9$ & $0.0309$ & $4$ \\
    OpenAI-O3 & $1.9$ & $-13.2$ & $0.0267$ & $5$ \\
    Qwen3-30B-Think & $2.1$ & $-13.5$ & $0.0255$ & $6$ \\
    Claude-4-Sonnet & $2.2$ & $-14.2$ & $0.0245$ & $7$ \\
    DeepSeek-V3.1 & $1.1$ & $-14.1$ & $0.0210$ & $8$ \\
    GPT-5 & $0.3$ & $-13.1$ & $0.0132$ & $9$ \\
    Qwen3-Coder & $0.2$ & $-13.9$ & $0.0137$ & $10$ \\
    DeepSeek-V3 & $0.2$ & $-14.1$ & $0.0144$ & $11$ \\
    Passive Baseline & $0.4$ & $-15.2$ & $0.0155$ & $12$ \\
    GPT-OSS-120B & $-0.9$ & $-14.0$ & $0.0156$ & $13$ \\
    GPT-OSS-20B & $-2.8$ & $-14.4$ & $-0.0069$ & $14$ \\
    \bottomrule
    \end{tabular}
    }
    \caption{
    The performance of tested models over the evaluation period. 
    The best performance in each metric is highlighted in bold. Models are ranked based on the z-score aggregation of all three metrics. RT stands for Final Return (\%), DDN stands for Max Drawdown (\%).
    }
    \label{tab:rank_percent_bold}
\end{table}

\section{Main Experiments}
\label{sec:exp}
In this section, we present the experimental setup and results of evaluating various LLM agents within the \ourbench trading workflow. 
We describe the trading environment, selected models, baseline strategy, and evaluation metrics.
We then analyze performance outcomes, highlighting key insights into the capabilities of LLM agents in real-world financial markets.

\subsection{Experiment Setup}
We detail the experimental setup for evaluating LLM agents in the \ourbench trading workflow.  
Specifically, we describe the trading environment, the models selected for benchmarking, the passive baseline, and the evaluation metrics used to assess performance.

\paragraph{Trading Environment.}   
The top 20 DJIA stocks are selected as the investment targets, ensuring diverse representation across sectors.
The evaluation period spans four months, from March 3 to June 30, 2025, covering 82 trading days and capturing a range of market conditions.
Each model starts with \$$100,000$ in cash and zero holdings, making daily trading decisions at market open.
Key inputs include (1) the historical actions on held stocks over the past seven days, (2) up to five recent news articles from the previous 48 hours, and (3) for selected stocks, fundamental data such as market capitalization, P/E ratio, dividend yield, 52-week high/low, and recent quarterly dividends.

\paragraph{Models to Evaluate.}  
We benchmark a diverse set of LLMs, including both open-weight models such as Qwen3~\citep{qwen3}\footnote{Without special denote, the Qwen3 series in this papers refers to the 2507 variants}, DeepSeek~\citep{r1,deepseekv3}, Kimi-K2~\citep{kimi-k2},  GLM-4.5~\citep{glm4.5} and GPT-OSS~\citep{gpt-oss}, as well as closed-source APIs like OpenAI's O3~\citep{o3} and Anthropic's Claude-4-Sonnet~\citep{claude4}. This selection covers a range of architectures, sizes, and training methodologies to assess generality across different LLM designs. 
All models are equipped with $32,768$ token context windows and decoded with official recommended settings to ensure their performance is optimized for the task.
To hance a reliable result, each LLM agents would be run three times with different random seeds, and the average performance is reported.

\paragraph{Passive Baseline.}  
As a reference point, we implement a passive equal-weight buy-and-hold strategy that allocates the initial capital equally across all selected stocks at the start of the evaluation period and holds these positions unchanged until the end. This naive allocation is a widely accepted benchmark in portfolio research, reflecting passive index tracking behavior and providing a robust lower bound against which more sophisticated active strategies can be compared~\citep{demiguel2009optimal,duchin2009markowitz}.

\paragraph{Evaluation Metrics.}  
We adopt three widely used measures in financial analysis:  

\subparagraph{Final Return.}    
This metric measures overall profitability as the percentage change in portfolio value from the initial amount \(V_0\) to the final amount \(V_T\):   
\[
\text{Final Return} = \frac{V_T - V_0}{V_0} \tag{1}
\]
It directly reflects the portfolio's overall performance over the evaluation period and is a simple, widely used measure of investment profitability~\citep{bodie2014investments}.

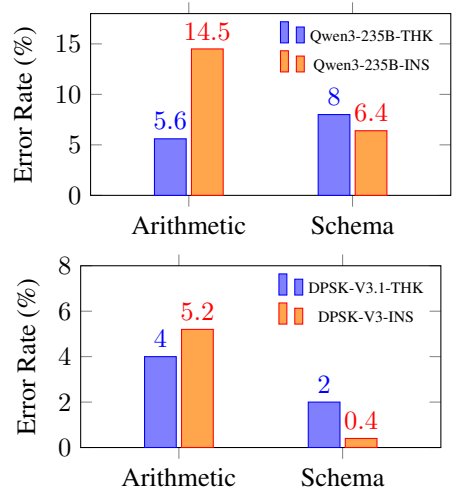
\begin{figure}[t]  
  \centering

  \begin{tikzpicture}
    \begin{axis}[
      ybar,
      bar width=12pt,
      width=\columnwidth,
      height=4cm,
      ymin=0,
      ymax=18,
      enlarge x limits=0.6,
      symbolic x coords={Arithmetic,Schema},
      xtick=data,
      ylabel={Error Rate (\%)},
      legend style={at={(0.99,0.99)}, anchor=north east, font=\tiny, draw=none, fill=none},
      nodes near coords,
      nodes near coords align={vertical},
    ]
      \addplot+[fill=blue!50] coordinates {(Arithmetic,5.6) (Schema,8.0)};
      \addplot+[fill=orange!70] coordinates {(Arithmetic,14.5) (Schema,6.4)};
      \legend{Qwen3-235B-THK, Qwen3-235B-INS}
    \end{axis}
  \end{tikzpicture}

  \vspace{0.25em}

  \begin{tikzpicture}
    \begin{axis}[
      ybar,
      bar width=12pt,
      width=\columnwidth,
      height=4cm,
      ymin=0,
      ymax=8,
      enlarge x limits=0.6,
      symbolic x coords={Arithmetic,Schema},
      xtick=data,
      ylabel={Error Rate (\%)},
      legend style={at={(0.99,0.99)}, anchor=north east, font=\tiny, draw=none, fill=none},
      nodes near coords,
      nodes near coords align={vertical},
    ]
      \addplot+[fill=blue!50] coordinates {(Arithmetic,4.0) (Schema,2.0)};
      \addplot+[fill=orange!70] coordinates {(Arithmetic,5.2) (Schema,0.4)};
      \legend{DPSK-V3.1-THK, DPSK-V3-INS}
    \end{axis}
  \end{tikzpicture}

  \vspace{-0.25em}
  \caption{Error distribution (\%) by type for Think vs Instruct models.}
  \label{fig:error_fig}
\end{figure}

\subparagraph{Maximum Drawdown.}  
The maximum drawdown quantifies the largest decline in portfolio value from its peak to its trough throughout the evaluation period, providing a measure for downside risk:  
\[
\text{Max Drawdown} = \min_{t \in [0, T]} \left( \frac{V_t - \max_{s \leq t} V_s}{\max_{s \leq t} V_s} \right) \tag{2}
\]
It highlights the worst loss an investor could have faced and is commonly used to assess risk and volatility~\citep{magdon2004maximum,chekhlov2005drawdown}.

\subparagraph{Sortino Ratio.}   
The Sortino ratio is a risk adjusted return metric that penalizes only downside volatility. It is defined as the excess return \(R_p\) divided by the downside deviation \(\sigma_d\):  
\[
\text{Sortino Ratio}=\frac{R_p}{\sigma_d},\quad
\sigma_d=\sqrt{\frac{1}{N_d}\!\sum_{i=1}^{N_d}\!\min(R_i,0)^2}\tag{3}
\]
This metric is more appropriate than the Sharpe ratio when returns are asymmetric, as it focuses on negative volatility~\citep{sortino1991downside,pedersen2002foundation}.

After computing these metrics for each model, we derive a composite rank by leveraging the z-score of each metric, averaging them to produce a single performance score. 
\[
\text{Composite Rank}
= \frac{z(\mathrm{Ret}) - z(\mathrm{DD}) + z(\mathrm{SR})}{3}
\tag{4}
\]

Where $\mathrm{Ret}$ is Final Return, $\mathrm{DD}$ is Max Drawdown, and $\mathrm{SR}$ is Sordino Ratio.
This approach balances profitability and risk, rewarding models that achieve high returns while effectively managing downside exposure.

\subsection{Experiment Results}
Table~\ref{tab:rank_percent_bold} presents the performance of all evaluated models over the four-month period without contamination. 
The results are reported across three key metrics---percentage return, maximum drawdown, and Sortino ratio---along with an overall ranking derived from a composite z-score of these metrics.

\begin{table}[!t]  
  \centering
  \vspace{0.065in}
  \setlength{\tabcolsep}{4pt} 
  \renewcommand{\arraystretch}{1.0} 
  \begin{tabular}{crrc}
    \toprule
    {Stocks} & {$\%$ Mean} & {$\%$ Std} & {CV} \\
    \midrule
    \multicolumn{4}{l}{\textcolor{gray}{\textbf{\textit{Kimi-K2}}}} \\
    $5$  & $-4.6$ & $0.7$ & $0.2$ \\
    $10$ & $3.2$  & $0.6$ & $0.2$ \\
    $20$ & $1.9$  & $1.7$ & $0.9$ \\
    $30$ & $-0.5$ & $1.2$ & $2.2$ \\
    \midrule
    \multicolumn{4}{l}{\textcolor{gray}{\textbf{\textit{GPT-OSS-120B}}}} \\
    $5$  & $-5.7$ & $0.3$ & $0.1$ \\
    $10$ & $2.5$  & $0.4$ & $0.2$ \\
    $20$ & $-0.4$ & $3.9$ & $10.2$ \\
    $30$ & $-0.9$ & $3.9$ & $4.4$ \\
    \bottomrule
  \end{tabular}
    \caption{Performance of representative models (Kimi-K2 and GPT-OSS-120B) across different investment target sizes. Results are reported as mean return (\% Mean), standard deviation of returns (\% Std), and coefficient of variation (CV).}
  \label{tab:stock}
\end{table}

Here are the key observations:
\textbf{(1) LLM agents can trade profitably in real-world markets.} 
Most tested models outperform the passive buy-and-hold baseline, which achieves a modest $0.4\%$ return with a $-15.2\%$ drawdown and a Sortino ratio of $0.0155$. Several agents deliver returns above $2\%$, with improved risk profiles.
\textbf{(2) LLM agents can manage downside risk effectively.} 
All tested models achieve lower maximum drawdowns than the baseline, indicating that they can mitigate losses during market downturns. The best-performing agents limit drawdowns to around $-11\%$ to $-14\%$, compared to the baseline’s $-15.2\%$.
\textbf{(3) Reasoning model does not guarantee better performance.} Although reasoning-tuned models such as Qwen3-235B-Think and Qwen3-30B-Think exhibit strong performance in tasks requiring complex reasoning, including math and coding~\citep{qwen3}, they do not consistently outperform instruction-tuned counterparts in this trading task. For example, Qwen3-235B-Ins outperforms its reasoning-tuned version with a lower maximum drawdown ($-11.2\%$ vs. $-14.9\%$).
This suggests there is still a gap between reasoning ability and effective decision-making in dynamic, noisy environments like financial markets.

\section{Analysis}
\subsection{Influence of Investment Target Size}
\label{subsec:scale}

To evaluate the impact of the investment target size on the agent’s performance, 
we study the effect of investment target size by running daily trading tasks on portfolios of 5, 10, 20, and 30 DJIA stocks, repeating each setting three times and measuring return variability.
The results show that variability increases as the investment target expands.
Specifically, as shown in Table~\ref{tab:stock}, 
\textbf{(1) Scalability is inherently challenging.} All evaluated models exhibit performance degradation as the investment portfolio size increases, characterized by declining mean returns and rising return volatility. This indicates that scaling the number of tradable assets poses a non-trivial challenge for LLM agents.
\textbf{(2) Model scale confers robustness.} The larger-scale model, Kimi-K2, demonstrates greater robustness to portfolio expansion, maintaining relatively stable risk-return profiles and achieving positive expected returns at moderate portfolio sizes (e.g., 10–20 stocks), whereas the smaller GPT-OSS-120B suffers from severe performance deterioration and excessive variability, suggesting that increased model capacity enhances generalization and stability in multi-asset decision-making contexts.

\subsection{Influence of Error in the Trading Workflow}

During trading, two common error types arise:  \textbf{(1) Arithmetic Errors}, where agents miscalculate share quantities. \textbf{(2) Schema Errors}, where the LLM agent's outputs violate the required JSON format. 
Figure~\ref{fig:error_fig} illustrates the frequency of these errors across thinking models and instruct  models.
The results show that thinking models make fewer arithmetic errors than instruct models, consistent with their stronger reasoning ability~\citep{yu2025dapo,r1,qwen3}.
Yet, they also incur more schema errors, likely due to their tendency to generate overly complex outputs that deviate from the expected format~\citep{fu2025scaling,li2025thinking}.



\subsection{Ablation Study on Data Sources}

\begin{table}[t]   
  \centering
  \setlength{\tabcolsep}{6pt}
  \renewcommand{\arraystretch}{0.9}
  \begin{tabular}{lr}
    \toprule
    Condition & Return (\%) \\
    \midrule
    \textbf{\textit{Kimi-K2}} & $1.9$ \\
    ~~\textit{w/o} News & $1.4$ \\
    ~~\textit{w/o} News \& Fund. & $0.6$ \\
    \midrule
    \textbf{\textit{GPT-OSS-120B}} & $-1.2$ \\
    ~~\textit{w/o} News & $-1.2$ \\
    ~~\textit{w/o} News \& Fund. & $-3.4$ \\
    \bottomrule
  \end{tabular}
   \caption{The cumulative return (CR, \%) for Kimi-K2 and GPT-OSS-120B 
  under three input settings: full input (Full), without news articles (w/o News), and without both news and fundamental data (w/o News \& Fund.).}
  \label{tab:ablation}
\end{table}


In our workflow, LLM agents primarily rely on two main information sources: news and fundamental data.
These two modalities provide complementary signals, with news capturing market sentiment and fundamentals grounding the model in key financial indicators.
To better understand their respective contributions, we conduct an ablation study by progressively removing these inputs. 
As shown in Table~\ref{tab:ablation}, cumulative returns drop as news and then fundamentals are removed, confirming their importance.
This behavior matches our expectation that both information sources play an important role in guiding trading decisions. 
The Kimi-K2 model remains relatively robust when only news is removed, but its performance deteriorates when both inputs are absent. 
In contrast, GPT-OSS-120B experiences a sharper decline, indicating that it relies more heavily on explicit signals provided by news and fundamentals. 
Overall, these results show that LLM agents effectively integrate textual and numerical inputs to guide trading decisions.



\subsection{Impact of Evaluation Window}

\begin{figure}[t]  
    \includegraphics[width=0.5\textwidth]{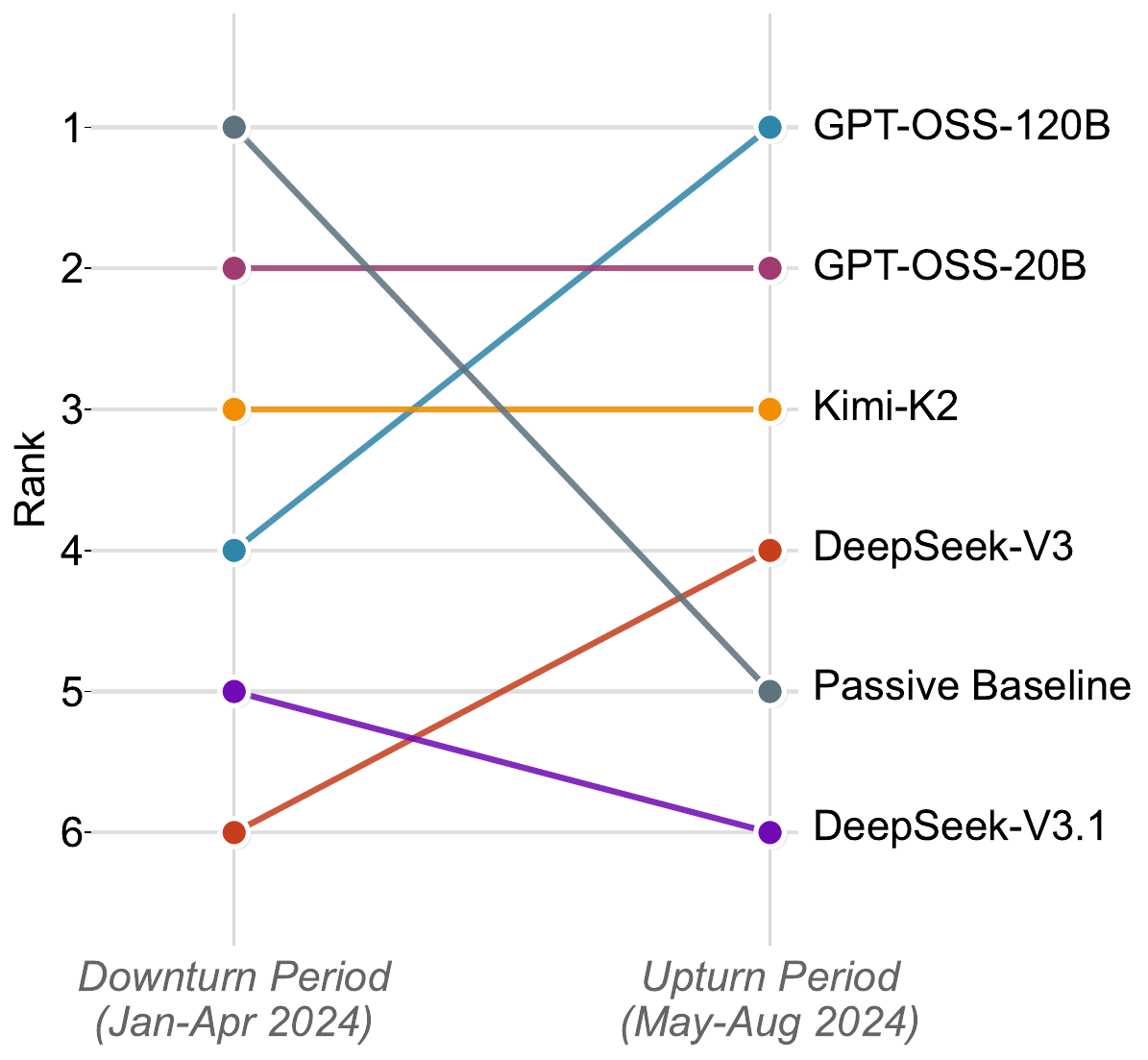}
    \caption{Model performance ranking based on the cumulative return,
    over two evaluation windows downturn (Jan-Apr 2025) and upturn (May-Aug 2025).}
    \label{fig:placeholder}
\end{figure}

A good trading model should be able to adapt to changing market conditions over time.
To investigate how the choice of evaluation window affects model rankings, we evaluate model robustness across two market time frames: a downturn period (January to April 2025) and an upturn period (May to August 2025). 
Figure~\ref{fig:placeholder} shows the models' ranking shifts between these periods. 
Notably, we observe significant shifts in model rankings between the downturn and upturn periods.
For example, GPT-OSS-120B moves from bottom-ranked in the downturn to top-ranked in the upturn, suggesting sensitivity to bullish markets, while Kimi-K2 remains relatively stable, indicating greater robustness to market fluctuations.
This suggests that certain models may be better suited to specific market conditions, potentially due to their underlying architectures or training data.
Notably, all LLM agents underperform the passive baseline during the downturn but outperform it in the upturn, highlighting a key weakness in bearish markets.



\section{Related Work}
\label{sec:related}

\subsection{LLM Agents and General Benchmarks}

Large language models (LLMs) have evolved from text generators into autonomous agents capable of reasoning, planning, and interacting with environments~\citep{gpt4o,claude3.7,gemini2.5,deepseekv3,r1,llama4,qwen2.5,qwen2.5vl,o3}.
Agentic behavior has been increasingly viewed as the next stage of LLM development, as it directly translates into real-world productivity~\citep{gpt5,claude4}. 
To evaluate these emerging capabilities, a range of agent benchmarks has been proposed across domains, including software engineering (SWE-Bench, SWE-Agent~\citep{jimenez2024swebench,yang2024sweagent}), scientific discovery (AstaBench~\citep{astabench}), and commercial workflows (XBench, Tau2Bench~\citep{chen2025xbench,barres2025tau2bench}). While these benchmarks demonstrate the potential of LLM agents for complex tasks, few existing works have examined domains where decisions translate into direct economic outcomes, such as financial trading.



\subsection{Financial Agents and Benchmarks}


Financial applications of LLMs have attracted growing attention due to their 
relevance to 
profitability, risk management, and high-stakes decision making~\citep{wu2023bloomberggpt,lee2024finllms,nie2024survey}. 
However, most existing benchmarks focus on static question-answering, such as FinQA, TAT-QA, and FinBench~\citep{chen2021finqa,zhu-etal-2021-tat,yin2023finbench}.
While useful for evaluating financial reasoning and domain knowledge, these benchmarks do not reflect the iterative, dynamic nature of real-world trading environments.
Recent works like INVESTORBENCH~\citep{li-etal-2025-investorbench} have begun to explore agent-based trading evaluation, but
primarily considers single-stock settings and relies on historical data up to 2021, raising concerns about both scope and potential data contamination.
In contrast, \ourbench is the first benchmark to embed LLM agents in realistic, multi-stock trading environments with continuously updated market data.
\ourbench bridges the gap between static financial QA benchmarks and the practical challenges of real-world investment strategies, enabling a more faithful assessment of LLM-based financial agents.

\section{Conclusion}

We introduce \ourbench, a benchmark for evaluating LLM agents in realistic stock-trading environments with dynamic markets and long-horizon decision making. 
Our experiments show that although current agents can be profitable, they rarely outperform simple baselines, revealing substantial room for improvement. We release \ourbench to support future research on building more capable trading agents under complex market dynamics.
We believe that \ourbench will serve as a valuable resource for the research community, driving further advancements in the development of intelligent, autonomous financial agents capable of navigating complex market dynamics. 


\section{Limitations}
\ourbench evaluates LLM agents using daily trading over several months on a fixed set of large DJIA stocks. While this setup is realistic and avoids data leakage, it does not cover high-frequency trading, long-term market cycles, or a wide range of assets. 
Since agents can only trade once per day, intraday and event-driven strategies are not tested. 
The four-month window may also miss long bull or bear markets and rare but important market events, which can affect model rankings. 
In addition, we focus on large U.S. stocks and do not model trading costs, slippage, or liquidity limits, which makes trading easier than in real markets. 
We view these limitations as important directions for future work, including extending \ourbench to longer horizons, higher-frequency trading, and more diverse asset classes to provide a more comprehensive assessment of LLM-based financial agents.

\section{Ethical Statement}
We strictly comply with all applicable financial regulations, data-protection laws, and academic ethical standards during the construction and use of \ourbench. All market data (prices, fundamentals, and news) were collected through licensed data vendors or public APIs that explicitly allow research use; no non-public, insider, or personally identifiable information was accessed or stored.
The benchmark is provided for academic and non-commercial research purposes only. Users are reminded that \ourbench is not intended to offer, or serve as the basis for, any financial advice, trading recommendation, or commercial activity. Any trading strategy tested on \ourbench carries inherent market risk; past performance recorded in the benchmark does not guarantee future returns.


\bibliography{acl_2026}

\appendix



\section{Prevent Data Leakage}
In this study, we minimize the risk of data leakage by carefully planning and evaluating the time frame. When testing large language models (LLMS) in the financial field, a potential concern is that during the training process, the model will learn a lot of past financial knowledge, which may lead to the model's performance being artificially exaggerated. For instance, when asking GPT-5 (without using the search function), we found that the model could accurately predict the stock trend of AAPL in 2021, and the model's response was consistent with the facts.

This discovery indicates that if the evaluation time is relatively early, the model may have obtained future information that could not have been reasonably acquired at the time of evaluation. In view of this, we have decided to limit the data used for evaluation to a more recent time frame, thereby minimizing the possibility of such ``data leakage'' and ensuring that the model is tested more fairly. By focusing on a narrow evaluation time window, we aim to simulate real-world scenarios where agents can only make trading decisions based on the publicly available information at the time of each decision.

This approach conforms to the best practices of financial model evaluation, ensuring that the evaluation results truly reflect the predictive and decision-making capabilities of LLM agents without being disturbed by the unintentional availability of future data

\section{Model Return Variance }


\begin{table}[h]
\centering
\caption{Model Return Variance Across Different Models. This table presents the variance of model returns for various LLMs.}
\label{tab:return-variance}
\begin{tabular}{lcrr}
\toprule
{Rank} & {Model} & {Var ($\times 10^{-4}$)} \\
\midrule
1 & \textit{DeepSeek-V3} & 0.074 \\
2 & \textit{DeepSeek-V3.1} & 0.203 \\
3 & \textit{GPT-5} & 0.210 \\
4 & \textit{Claude-4-Sonnet} & 0.153 \\
5 & \textit{GLM-4.5} & 0.099 \\
6 & \textit{Qwen3-30B-Think} & 0.115 \\
7 & \textit{Qwen3-235B-Think} & 0.321 \\
8 & \textit{Qwen3-235B-Ins} & 0.281 \\
9 & \textit{Qwen3-4B-Ins} & 1.382 \\
10 & \textit{GPT-OSS-20B} & 1.337 \\
11 & \textit{Qwen3-Coder} & 1.655 \\
12 & \textit{Openai-O3} & 3.250 \\
13 & \textit{Kimi-K2} & 1.866 \\
14 & \textit{GPT-OSS-120B} & 10.19 \\
\bottomrule
\end{tabular}
\end{table}

In this section, we analyze the return variances of different models. Models with higher return variances may exhibit more unpredictable behaviors, which is undesirable in many real-world applications, especially in high-risk environments such as financial decision-making.

We ranked several large language models (LLMS) based on their return variances, as shown in table \ref{tab:return-variance}. In the evaluated model, \textit{DeepSeek-V3} exhibited the smallest performance fluctuation, indicating high stability. In contrast, \textit{GPT-OSS-120B} exhibits the highest return variance, indicating a volatility in its performance.





\section{The Use of Large Language Models}

We use LLMs for two purposes.
(1) Code implementation. When implementing the code for this paper, including data gathering and experiment implementation, we use LLMs in the form of \texttt{copilot} to complete code snippets.
The architecture design is conducted by human researchers.
(2) Proofreading. To fix grammar issues, we use LLMs as a writing tools to refine the draft.

We would like to highlight that LLMs are not responsible for creativity tasks during conducting the research of this paper, including but not limited to: ideation, experiment design, paper organizing.

\end{document}